\documentclass[twoside,11pt]{article}

\usepackage{graphicx}
\usepackage{epstopdf}
\usepackage{url}
\usepackage{amssymb}
\usepackage{amsmath}
\usepackage[margin=2.5cm]{geometry}

\renewcommand{\a}{\mathbf{a}} 
\renewcommand{\o}{\mathbf{o}} 
\renewcommand{\b}{\mathbf{b}}

\renewcommand{\P}{\mathbf{P}} 
\newcommand{\D}{\mathbf{D}} 
\newcommand{\U}{\mathbf{U}} 
\newcommand{\W}{\mathbf{W}} 
\renewcommand{\r}{\mathbf{r}} 
\newcommand{\e}{\mathbf{e}}

\newcommand{\h}{\mathbf{h}}

\newcommand{\g}{\mathbf{g}}

\begin{document}
\title{Towards Trainable Media: \\Using Waves for Neural Network-style Learning}

\author{Michiel Hermans\thanks{michiel.hermans@ulb.ac.be Laboratoire d'Information Quantique, Universit\'e Libre de Bruxelles (U.L.B.), 50 avenue F.D. Roosevelt, CP 225, B-1050 Brussels, Belgium}, Thomas Van Vaerenbergh\thanks{Photonics Research Lab, INTEC department, Ghent University, Sint Pietersnieuwstraat 41, 9000 Ghent, Belgium. Current address: Hewlett Packard Labs, 1501 Page Mill Road, Palo Alto, CA 94304, United States}}

%\keywords{Keyword1, Keyword2, Keyword3}

\flushbottom
\maketitle

\thispagestyle{empty}

\noindent 
\begin{abstract}
In this paper we study the concept of using the interaction between waves and a trainable medium in order to construct a matrix-vector multiplier. In particular we study such a device in the context of the backpropagation algorithm, which is commonly used for training neural networks. Here, the weights of the connections between neurons are trained by multiplying a `forward' signal with a backwards propagating `error' signal. We show that this concept can be extended to trainable media, where the gradient for the local wave number is given by multiplying signal waves and error waves. We provide a numerical example of such a system with waves traveling freely in a trainable medium, and we discuss a potential way to build such a device in an integrated photonics chip.
\end{abstract}

\begin{figure}[t]
\begin{center}
\includegraphics[width=0.7\textwidth]{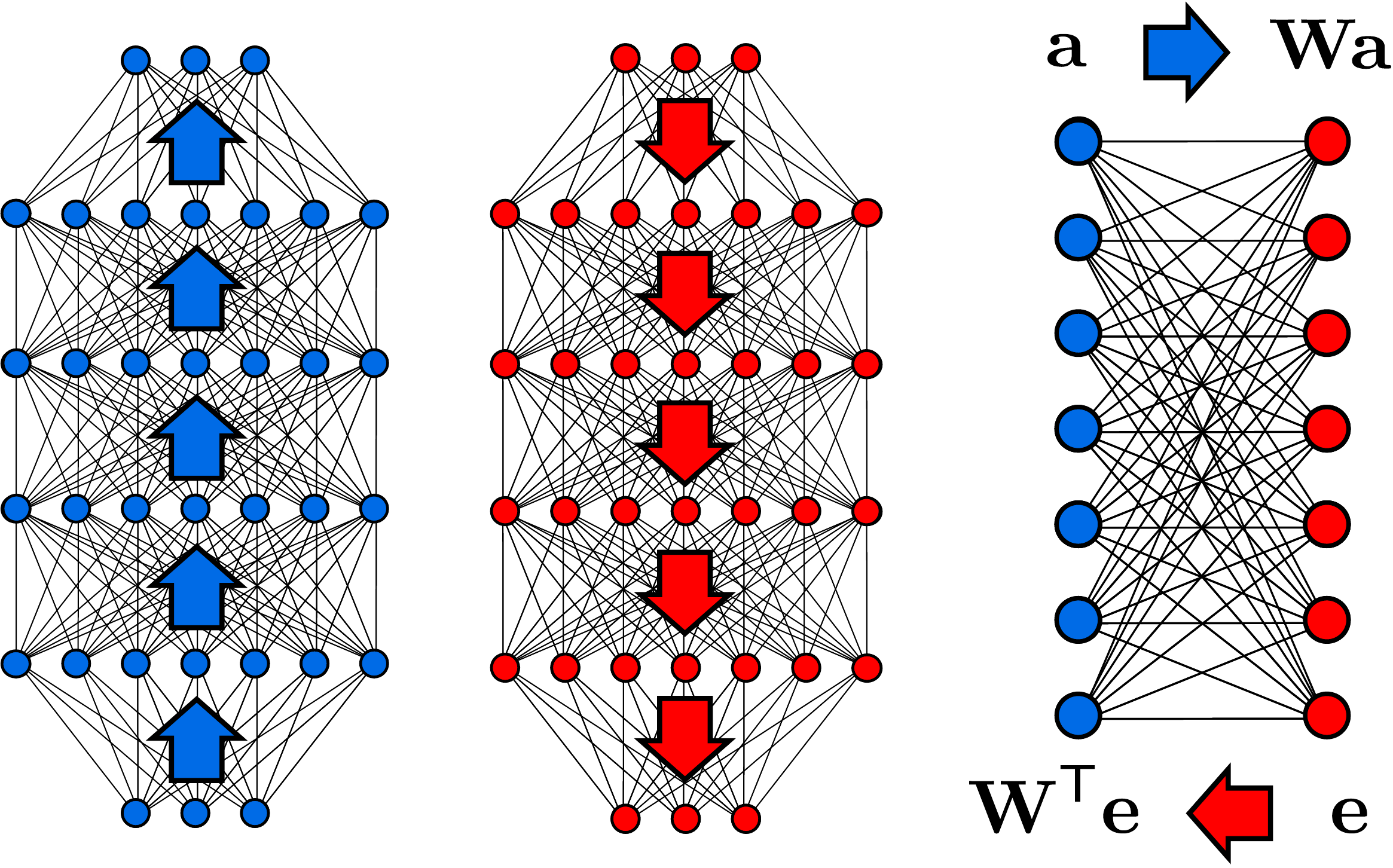}
\caption{Schematic representation of a multi-layered neural network. Each neuron (circle) receives a weighted sum of of the states of the previous layer and performs a non-linear transformation. Commonly, input is presented at the bottom layer, and output at the top, and information flows from bottom to top. During the training phase, the error at the top layer (e.g., the difference between the actual and desired output for a given input instance) is backpropagated through the network over the same connections, and in each neuron multiplied with the derivative of the nonlinearity. Finally, the weighted connection are updated in each layer by multiplying the forward signal on one side with the error signal at the other. On the right we depicted a basic trainable block (not including the nonlinearity), which is able to multiply an input vector $\a$ with a matrix $\W$, multiply an error $\e$ with $\W^\textsf{T}$, and is able to adapt its weights based on $\a$ and $\e$.}\label{fig:NN}
\end{center}
\end{figure} 
The last decade has been marked by a conspicuous renaissance of neural network research and development. A field of study which was once confined largely to theory, has bloomed into a highly applied research domain, which succeeds in tackling intricate problems such as speech recognition \cite{Hinton2012} and computer vision \cite{Krizhevsky2012} with unrivalled success. The main driving forces behind this development are the more widespread availability of cheap, high-powered digital processors, the advent of broad-purpose massively parallel computing on GPUs, and the availability of very large datasets. The combination of these allowed researchers to drastically scale up and accelerate their models, in turn allowing an exploration of the limits of their performance. Currently, neural networks (NNs) outperform all other approaches in a variety of applications \cite{LeCun2015}. Trends include the exploration of very challenging linguistic problems such as machine translation \cite{Sutskever2014, Cho2014}, and the integration of several types of tasks into single systems such as the automatic captioning of pictures \cite{Socher2014} and training agents to play arcade games \cite{Mnih2013}. NNs are remarkably successful in these tasks, and the limits of their capabilities seem yet far from being reached.\\
NNs are still trained using a straightforward, half-century old algorithm called backpropagation \cite{Rumelhart1986,Werbos1988}, which computes gradients for the parameters of the networks. A generic neural network structure and its associated training process is visualised in Figure \ref{fig:NN}, as well as a representation of a basic, trainable element.\\
By far the most computationally intensive part of NN implementations--both in training and during the actual operation--are the required matrix-vector multiplications, i.e., large-scale linear transformations. Increasing the speed and scale of such computations for digital processors always comes with a significant cost in terms of power consumption. Therefore a sizeable research effort has been made in the past to build matrix-vector-multipliers (MVM) using analog processors, where a linear transformation is performed through a physical process rather than by explicit digital computation. In this paper we aim not just at an MVM, but rather at constructing a basic trainable neural network connection layer, as represented in Figure \ref{fig:NN}. This means it needs to have the following properties
\begin{itemize}
\item Bidirectionality: the ability to perform a multiplication with a matrix in one direction, and the transposed matrix in the other.
\item Trainability: the ability to adapt itself by combining information of the forward signal and the error signal. Ideally this should be a local operation, where no external processing is required.
\end{itemize}
Note that in this paper we are not concerned with adding a nonlinear function, (or a multiplication with the local derivative in the backpropagation phase), but rather purely focus on have a trainable linear transformation. If such a device can be constructed, a full neural network can be constructed by adding intermediate nonlinearities.
\begin{figure}[t]
\begin{center}
\includegraphics[width=0.8\textwidth]{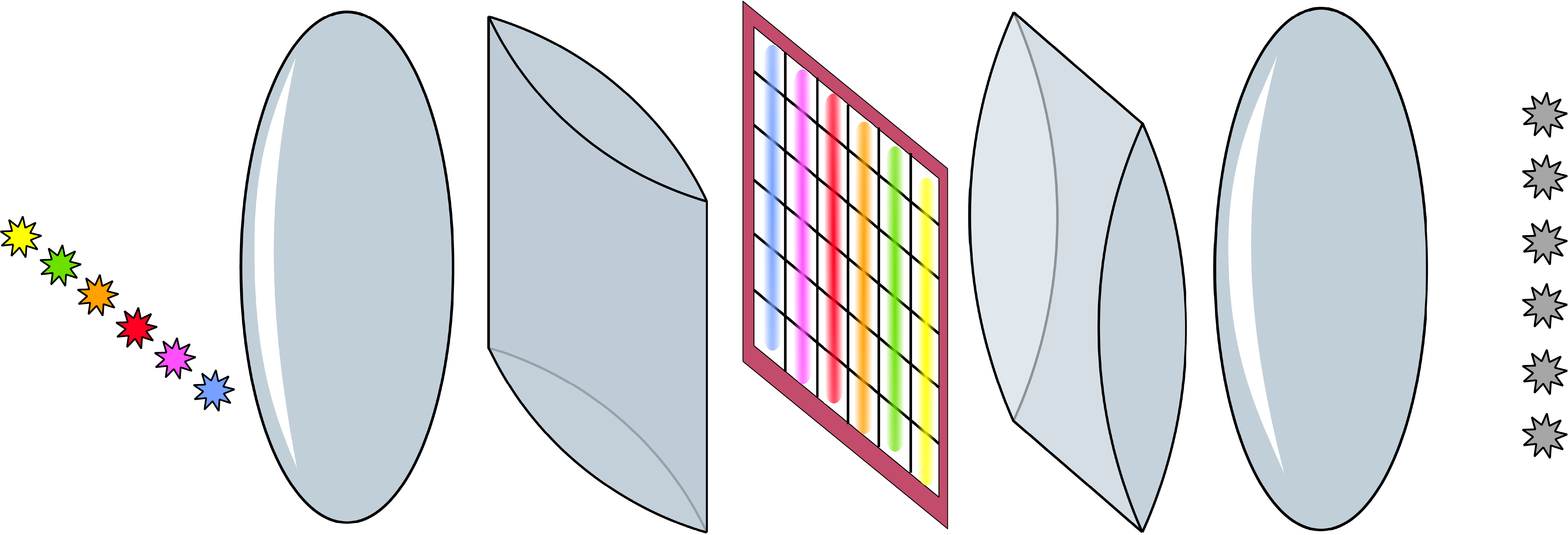}
\caption{Schematic representation of an example of a spatial light modulator used for matrix-vector multiplication. On the left we have six light sources that encode a vector of numbers (either through intensity or complex amplitude depending on the kind of implementation). By a combination of a normal and cylindrical lens, each source is spread and focused on a single column of the spatial light modulator. Each transparent pixel modulates the light falling into it (either its intensity or phase, or both, depending on the specific device and implementation). The optical setup on the right side reverts this operation, and the light emitted by each individual row of the spatial light modulator is now focused onto one of six detectors on the far right. This operation essentially performs a row-wise sum of the light exiting the spatial light modulator. Note that we can exchange the locations of sources and receivers without changing the functionality of the device.}\label{fig:SLM}
\end{center}
\end{figure} 
Optically implemented MVMs have been studied extensively \cite{Farhat1985,Yeh1987,Goodman1978,Caulfield1989}, where very fast matrix-vector multiplications are performed by passing light through a spatial light modulator (e.g., a transparent LCD screen, see Figure \ref{fig:SLM}). Here, the input vector elements are encoded in an array of light sources (either as complex amplitudes or as light intensities) and the matrix elements are encoded as pixel values on a spatial light modulator. Such a device (and its many variants) works very well to perform vector matrix multiplications and it is in principle bidirectional, meaning that we simply can pass light through the system in the opposite direction in order to implement a transpose matrix multiplication. It is bulky, however, relying on free-space optics, and cannot easily be integrated into a compact device.\\
One integrated-optics MVM has been proposed \cite{Yang2013} where the input vector is encoded in light intensities at different wavelengths, and matrix elements are implemented by modulating the refractory indices of an array of microring resonators. While such a device is certainly compact, it is not sure how scalable it would be (the number of wavelengths that can be applied is typically limited by the properties of the waveguides and the ring resonators). Furthermore, as signal summation happens in the intensity domain, there is no straightforward way in which the transpose matrix can be implemented, though perhaps such a functionality could be added by changing the design.\\
A promising and more recent development is the use of arrays of memristor devices \cite{Deng2015, Prezioso2015}. Here, weights are stored as the individual conductances of a large array of memristors in a crosshair circuit. Such a device can be miniaturised easily and has properties that are very desirable for NN applications. While conductance is a strictly positive quantity, four-quadrant multiplication can be achieved using differential pairs of currents, such that a weight is encoded as a difference in conductance between two memristors. No existing implementations can yet perform the transpose matrix multiplication, but perhaps future models could incorporate such a functionality.\\
All the devices that are discussed above explicitly encode the matrix elements as a two-dimensional array of physical variables, closely adhering to the underlying structure of the problem. In this paper we will consider an alternative approach, where we drop the one-to-one correspondence to system parameters and matrix elements. Take for example the following situation. We have a room in which there are a set of acoustic speakers emitting sine waves, all with the same frequency, but with different amplitudes and phases, which can be expressed by a vector of complex amplitudes $\a$. On the other side of the room are a set of microphones. They will receive a set of signal which we can express as a vector of complex amplitudes $\b$. When we assume no nonlinear effects, the signals are simply linear combinations of the complex amplitudes $a_j$, i.e., $\b = \W\a$. The transition matrix $\W$ is a function of the shape of the room, the reflection of the walls, the positions of the speakers and microphones, etc.\\
Such a system is reciprocal; this means that we can multiply with the transpose of $\W$ by emitting a signal at the locations of the receivers, and measuring at the location of the original microphones (see \cite{Hermans2015} for more details). Generally, however, the relationship between the parameters that describe the properties of the room and the elements of $\W$ is difficult to model; it would be a hard task to find the shape of the room which would implement a given $\W$. The problem we will discuss in this paper is a similar one.  Rather than finding the shape of a room, we consider a planar structure in which the wave number $k(\r)$ can be varied locally: for example, the local refractory index of an optical planar wave-guide. The task then becomes: find a function $k(\r)$ that implements a given linear transformation between the set of emitters and receivers.\\
What we will argue in this paper, is that even though finding a set of parameters that directly implements a certain linear transformation $\W$ is too hard to be practical (especially in an iterative NN training setup), finding the \emph{gradient} of these parameters w.r.t. a certain cost function can be straightforward. This means that, if we would incorporate such a device in a physical analog implementation of a NN, we are able to \emph{train} it in a way which is very similar to how weight matrices are commonly trained in NNs. For this we will rely on properties of the wave equation. We will show that the gradient for a space-dependent parameter (particularly the wave number $k(\r)$) can be determined by correlating the waves propagating in the forward direction (which encode a signal), and waves propagating in the backwards direction (encoding an error signal). Concretely, we show that all the necessary information for determining a local parameter is present at the location where it is physically manifested. This means that we can adapt the parameters of such a device without needing to process information non-locally, potentially leading to very fast neural hardware.
\\
A number of devices based on neural networks and implemented in integrated photonics have been presented \cite{Tezak2015, Vandoorne2014}. In \cite{Miller2013,Miller2015} it is argued that one can construct a self-tuning optical device to perform a conversion from one set of orthogonal modes to another with only a minimal need for external computations. Similarly, in \cite{Mower2014,Steinbrecher2015} the concept of on-chip, fully tuneable optical transformations is considered, using an array of integrated Mach-Zehnder interferometers. Both these works use constructive methods to determine the tuneable parameters in the network, which is not applicable to the typical neural network setting that works with pairs of input-output examples.\\ 
The work presented in this paper is essentially a physically implemented version of adjoint optimisation, which is a standard optimisation strategy for complex systems in many industries. Indeed, similar methods as the ones presented in this paper can be used for numerical optimisation of optical components. Interestingly, there seems to be only little attention for such techniques in the photonics literature (some examples include \cite{Jensen2011,Seliger2006,Lalau2013,Liu2013,Niederberger2014}). It should be stated that they are in many ways superior to blind optimisation such as, e.g., the particle swarm algorithm \cite{Poli2007}, which is popular in nanophotonics design \cite{Robinson2004,Roelkens2008,Zhang2013}.\\
Using light itself for error backpropagation for neural network applications has been suggested and analysed before in a specific setup, where the Kerr nonlinearity is used to spatially modulate the refractory index of a thin layer of material \cite{Skinner1995}. Later the same group provided an experimental demonstration of this concept \cite{Steck1999}, which unfortunately received little attention. The advent of integrated photonics, and the concepts provided in this paper, will hopefully revive this line of research.
\section*{Results}
\subsection*{Gradient descent on the Helmholtz equation}
In this section we present the gradient of a certain cost function w.r.t. a certain parameter for a system described by the Helmholtz equation. Suppose we have a field $\phi_a(\r)$ which adheres to the following equation:
\begin{eqnarray}
\nabla^2\phi_a(\r) + k^2(\r)\phi_a(\r) &= &a(\r)\textrm{ for } \r\in\Omega\nonumber\\
\phi_a(\r) &= &0 \textrm{ for } \r\in\partial\Omega,
\end{eqnarray}
where $k(\r)$ is the local wave number, and $a(\r)$ is a certain source term which can for instance encode an input vector. We assume there exists a certain cost functional $Q(\phi_a)$ that we want to minimise by adapting $k(\r)$. It is possible to show (see the supplementary material) that the gradient of this cost functional w.r.t. $Q(\phi)$ is proportional to:
\begin{equation}g(\r) \sim -\Re(\phi_a(\r)\phi_e(\r)).\label{eq:grad2}\end{equation}
Here, $\phi_e(\r)$ is a second complex field (which can be interpreted as the `error' signal), which adheres to the following equation:
\begin{eqnarray}
\nabla^2\phi_e(\r) + k^2(\r)\phi_e(\r) &= &e(\r)\textrm{ for } \r\in\Omega\nonumber\\
\phi_e(\r) &= &0 \textrm{ for } \r\in\partial\Omega,\nonumber
\end{eqnarray}
with 
\begin{equation}
 e(\r) = \frac{\partial Q(\phi_a)}{\partial \phi_a(\r)}.
 \end{equation}
 This means that the information for the local gradient can be obtained by combining information from two complex fields, both of which are solutions of the wave equation, and therefore possible to generate physically.\\
We can now associate these equations with a trainable neural network block as follows. First of all, we assume that the source term $a(\r)$ emerges from an array of emitters, and can be related to an input vector $\a$ as follows:
\begin{equation}
a(\r) = \sum_{i=1}^{N_a} a_i \beta_i(\r),\label{eq:source_a}
\end{equation}
with $a_i$ the $i$-th element of $\a$ of size $N_a$, and $\beta_i(\r)$ a source field associated with the $i$-th emitter. Next we assume that there exists an array of $N_o$ receivers which make measurements the complex field $\phi_a(\r)$ to generate an output vector $\o$ of dimensionality $N_o$, where each element $o_i$ is described by:
\[
o_i = \int_\Omega \phi_a(\r)\gamma_i(\r) d\r,
\]
with $\gamma_i(\r)$ a function describing the properties of the $i$-th receiver. Here it is important to note that--due to the fully linear nature of the entire system--the output vector $\o$ can be written as $\o = \W\a$, i.e., there exists a matrix $\W$, determined by the combination of the wave number function $k(\r)$, the boundary conditions, and the emitter-receiver functions $\beta_i(\r)$ and $\gamma_i(\r)$. In this sense, the whole system acts as an implicit MVM.\\
In the common neural network training scheme, there will be a certain cost vector $\e$ associated with the output $\o$, which represents the gradient of an external cost function w.r.t. the individual elements of $\o$. We can relate this with the cost functional $Q(\phi_a)$ by explicitly writing the dependencies: $Q(\phi_a,\phi_a^*) = Q(\o(\phi_a),\o^*(\phi_a^*))$ (note that $\o$ by definition does not depend on $\phi_a^*$). Therefore, the functional derivative $e(\r) = \partial Q(\phi_a,\phi_a^*)/\partial \phi_a(\r)$ can be written as
\[
e(\r) = \sum_{i=1}^{N_o} \frac{\partial Q\left(\o(\phi_a),\o^*(\phi_a^*)\right)}{\partial o_i}\frac{\partial o_i}{\partial \phi_a(\r)} =  \sum_{i=1}^{N_o} e_i\gamma_i(r),
\] 
where $e_i$ are the individual elements of $\e$. Note that this equation takes on a very similar form as Equation \eqref{eq:source_a}. Indeed, when we can use the emitters and receivers in two directions i.e., when they act as transducers), it is possible to simply \emph{emit} the error signal into the system.
\subsection*{Numerical example of an MVM}
\begin{figure}[t]
\begin{center}
\includegraphics[width=0.95\textwidth]{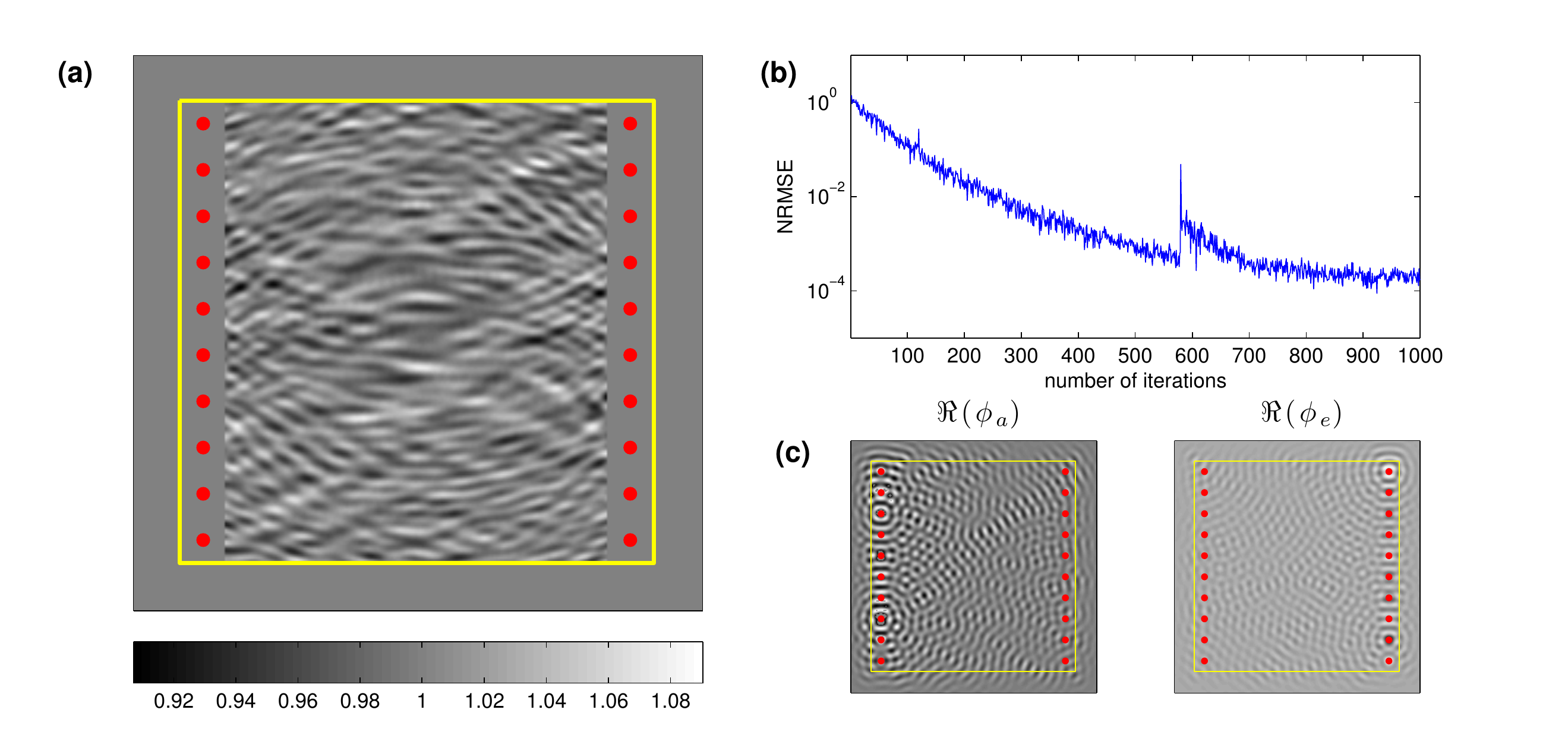}
\caption{Simulation results of the wave-based MVM. \textbf{(a)}: Depiction of the relative refractory index of the slab after optimisation. The rectangular area in the middle has a variable wave number. The colour bar at the bottom indicates the relative change from the average value. The red dots indicate the positions of the emitters/receivers. Outside the yellow boundary the medium becomes absorbing. \textbf{(b)}: The NRMSE shown as a function of the number of training iterations. \textbf{(c)} Example images showing the real part of $\phi_a$ and $\phi_e$, emitted from the left and right, respectively. }\label{fig:planar_results}
\end{center}
\end{figure}
In this section we will apply the previously derived concepts to a numerical examples. We will train a slab of material to perform a matrix-vector product of the input vector $\a$ with a pre-specified matrix $\W$. As we wish to use the classic neural network training mechanism, we will do this by drawing examples of input and desired output, and as a cost function we will use the squared error between the actual and desired output. Suppose we draw an input vector $\a$, with complex elements where the real and imaginary parts have been drawn from a standard normal distribution. Using the previously defined ideas, this is then sent into a medium encoded as waves, and converted back into an output vector $\o$. If we define $\o^t = \W\a$ as the ``target'' output, the cost function is given by
\[
Q(\o,\o^*) = ||\o- \o^t ||^2 = \sum_{i=1}^{N_o} (o_i - o^t_i)(o_i - o^t_i)^*
\]
We find that the derivative $e_i = \partial Q(\o,\o^*)/\partial o_i$ is given by 
\[
e_i = \frac{\partial Q(\o,\o^*)}{\partial o_i} = (o_i - o^t_i)^*.
\]
Note that the gradient found for a single input/output example pair will not suffice to make the system emulate $\W$. Rather, we will use an iterative optimisation scheme such that at the $k$-th iteration we update $k(\r)$ as follows:
\[
k(\r) \leftarrow k(\r) - \eta_k g_k(\r),
\]
with $\eta_k$ the learning rate, and $g_k(\r)$ the gradient obtained using input-output pair sample  $\a_k$ and $\o^t_k$. Such an optimisation scheme is the equivalent of stochastic gradient descent used in neural network training, where each training iteration only a small fraction of the data is presented to the network. In our case this is the extreme case of a single data point per iteration.\\
Figure \ref{fig:planar_results}(a) shows a schematic depiction of the simulated setup. Further details of the experiment are explained in the supplementary material. 
The results are displayed in Figure \ref{fig:planar_results}. The medium adapts itself successfully to perform the required linear transformation between input and output fields, as can be seen from Figure \ref{fig:planar_results}(b), which shows the evolution of the NRMSE as a function of the number of training iterations. The sudden peak at around the $580$-th iteration likely has to do with poor convergence of the method we used to compute the complex fields. We also found that the lowest NRMSE the system can reach during training depends on the precision tolerance used in this computation. The more precise the complex fields are computed, the lower the NRMSE gets.\\
Figure \ref{fig:planar_results}(a) depicts the structure of the simulated device, where we plotted the varying refractory index as a function of space. We have included the scaling of the colour coding, which shows the relative magnitude of the variations w.r.t. the average (equal to one). As can be seen, these variations remain relatively small, staying within 10 \% of the average value. Figure \ref{fig:planar_results}(c) shows examples of the complex fields of $\phi_a$ and $\phi_e$. 
\subsection*{Optical implementation}
Here we will detail the requirements to implement a trainable MVM using optics. We will assume the trainable medium consists of a planar material of which the refractory index can be modulated spatially, for example by using the Pockels effect in indium phosphide or lithium niobate, by using free carrier dispersion in semiconductors , or by the thermo-optic effect, present in many materials including silicon.\\
In many cases, the Maxwell equations in a two-dimensional structure can be reduced to a scalar wave equation, which makes optics a viable implementation platform for a trainable MVM.\\
Looking at the expression for the gradient (Equation \eqref{eq:grad2}), one would need to make a measurement of the real part of the product of two complex fields, over the entire region where one wishes to modulate the refractory index. This requires a system-wide interferometric measurement of the complex fields $\phi_a$ and $\phi_e$, such that we can directly measure their respective real and imaginary parts. In a planar material, this is--at least in principle--possible, as we could access the evanescent field at any location. In reality, however, this would pose a significant challenge, as for interferometry we would need to have access to a reference signal at every location too. Even when we succeed in separately measuring $\Re(\phi_a)$, $\Im(\phi_a)$, $\Re(\phi_e)$, and $\Im(\phi_e)$, we would still need to compute the quantity 
\[\Re(\phi_a\phi_e) = \Re(\phi_a)\Re(\phi_e) - \Im(\phi_a) \Im(\phi_e)\]
for each location, greatly increasing the device's complexity.\\
An alternative way of obtaining the gradient is to generate the complex conjugate field of either $\phi_a$ or $\phi_e$. If we are capable to do so, we could make three system-wide field intensity measurements (which are far simpler than interferometric measurements) as follows:
\begin{eqnarray}
2\Re(\phi_a\phi_e) &=& \phi_a\phi_e + \phi_a^*\phi_e^*\nonumber \\
&=& (\phi_a + \phi_e^*)(\phi_a^* + \phi_e) - \phi_a\phi_a^* - \phi_e\phi_e^*\nonumber\\
&=&I(\phi_a + \phi_e^*) - I(\phi_a) - I(\phi_e)\label{eq:grad3},
\end{eqnarray}
with $I(\phi)$ indicating the intensity of the field $\phi$. Immediately obvious from this expression is that the necessary computations are far simpler (additions instead of multiplications, which can be performed in an analog fashion easily). Indeed, using this approach seems more promising than direct measurements of $\phi_a$ and $\phi_e$. The issue then shifts to the following question: how well are we able to generate the field $\phi_a^*$ (or $\phi_e^*$)?\\
Generating the complex conjugate of an optical field is a well-studied problem, and has many potential applications in microscopy, lasers, sensors, holography, etc. Especially the concept of complex conjugate mirrors to correct phase distortions has received lots of attention. In our case, Equation \eqref{eq:grad3} shows that we don't require a complex conjugate mirror, but rather, we'd need to generate the complex conjugate field separately (at a different instant than which the original field is present) and add this to another optical field. One possible approach to achieve this would be to construct an optical phased array. If we could measure the phase front $\phi_a(\r_f)$, with $\r_f$ consisting of the set of points where light exits the system, \emph{and} we are able to regenerate the complex conjugate of this phase front as a source, we can--in principle--generate $\phi_a^*$ and $\phi_e^*$. There is one important caveat though: there should be no significant internal losses anywhere within the medium; phase conjugation can only invert phase distortions but not optical losses. This means that we would need to capture and detect \emph{all}  the light that exits the system. Internal losses are unfortunately ubiquitous in integrated photonics. In our case they are even a necessity if we want to measure the optical intensity throughout the trainable medium. 
\subsection*{Trainable waveguides}
\begin{figure}[t]
\begin{center}
\includegraphics[width=0.9\textwidth]{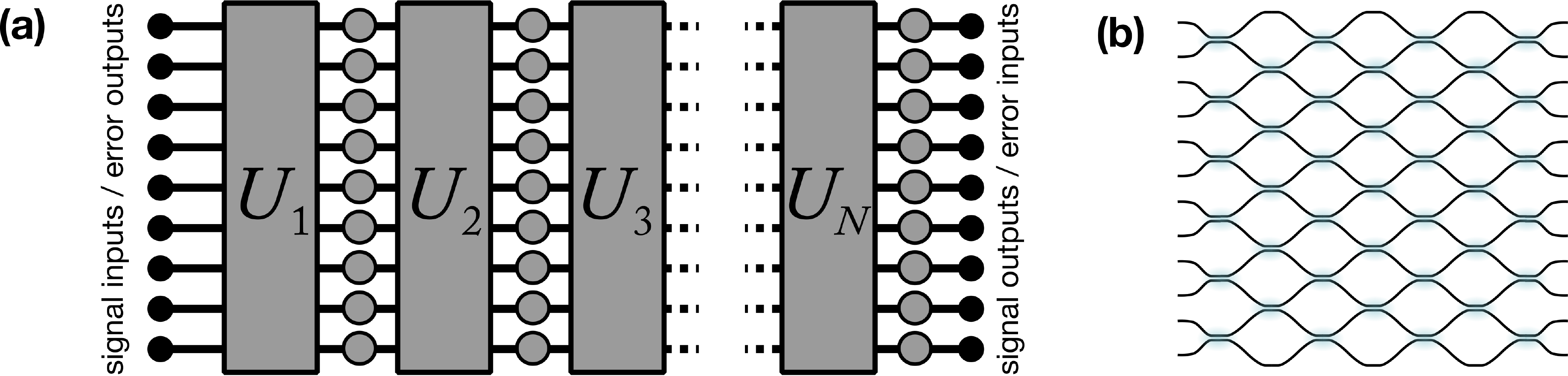}
\caption{Structure of a trainable nanophotonic chip. \textbf{(a)}: Possible design of a trainable optical circuit. The black lines are waveguides. The grey boxes represent unitary transformations of the optical signals, and the grey circles represent tuneable phase shifters. \textbf{(b)}: Example of a waveguide structure which would implement a `useful'  unitary transformation. The places where the waveguides approach each other very closely are directional couplers which allow light to be exchanged between the waveguides without losses. This structure was the one used for the numeric simulations, though in reality, since the only role of the unitary matrices is to `mix' the light form different waveguides, likely more compact and convenient designs are possible.}\label{fig:chip}
\end{center}
\end{figure}
Clearly, bulk material slabs in which light can propagate freely may pose too many challenges. We need an optical system that offers more control over the complex field that is generated within. For this, let us consider optical circuits that consist of single-mode waveguides, which greatly simplifies the conceptual description of the optical field; the optical signal present in each waveguide can be described by a single complex number (when considering one direction of travel). Note that optical circuits simply are defined by the refractory index of the system varying in space. This means that the theory that has been presented in the first part of this paper can be applied to such systems too.\\ 
Let us consider the aforementioned theory on an ideal waveguide. Equation \eqref{eq:grad3} tells us how we can express the gradient w.r.t. the local refractory index as a function of intensities. In the case of a waveguide, $\phi_a(\r)$ can represent a guided mode with light traveling in one directional mode which we could consider the `forward' direction. The field $\phi_e(\r)$ will represent a guided mode traveling in the opposite direction, i.e., `backwards'. Its complex conjugate, $\phi_e^*(\r)$ is reversed in direction, and therefore corresponds to a guided mode traveling also traveling forward, in the same direction as $\phi_a(\r)$. If we consider the intensity $I(\phi_a(\r) + \phi_e^*(\r))$, throughout the waveguide, this is the intensity of a guided mode, unchanging in the direction of travel. Therefore, the gradient one obtains within a single  waveguide is uniform throughout its length. It follows that the primary effect of adapting the refractory index will be to change the phase shift induced by this waveguide. It also follows that it suffices to measure light intensities at a single location within the waveguide. Tuneable phase shifters are an integral part of integrated photonics, and can be implemented in several ways (mechanically, electro-optically, or thermo-optically).
\subsubsection*{Trainable optical circuit}
We start from the idea that we have an optical chip with $N_a$ input waveguides and $N_a$ output waveguides. Using trainable waveguides as phase modulators, we then consider a structure as represented in Figure \ref{fig:chip}(a). Here, the signal alternates between being passed through a set of tuneable waveguides (phase shifters) and being transformed with fixed, unitary transformations $\U_i$.\\ 
The presented structure would need to have coherent detection and generation of signals exclusively at the in- and output sources. Generating the conjugate field $\phi_a^*$ or $\phi_e^*$ would only require the coherent measurement of the field at the receivers (which is a requirement in any case to produce the output vector $\o$), and generating the complex conjugate to be sent back into the system. After this, we need to perform the three measurements mentioned before and measure the light intensities from Equation \eqref{eq:grad3} within each waveguide in order to know how to change its refractory index.\\
To shed some light on the functionality of such a device and foreseeable issues, we will simulate its training process under successively less ideal circumstances. Note that the device represented in Figure \ref{fig:chip} will (if there are no losses) always perform a unitary transformation from input to output. Therefore, we will use a unitary matrix as a target for training.\\
In the supplementary material we explain in detail how the chip is modelled, and how the training is performed. During a single training update, `signal' light enters the chip from the left, is transformed by the chip, and is coherently detected at the output (right side). In order to perform the training, we generate the complex conjugate of this light at the right side, add it up to the `error' input, and send it backwards into the chip such that we can detect $I(\hat{\phi}_a^* + \phi_e)$, where $\hat{\phi}_a^*$ stands for the approximation of the complex conjugate field $\phi_a^*$ (the difference is due to losses in the system). Next we send in the error signal and the conjugated output signal separately to determine $I(\hat{\phi}_a)$ and $I(\phi_e)$. In total this means that three measurements are necessary to obtain the gradient, all based on light propagating through the chip in the backwards direction. Alternatively, we can work from the left side, where we use an approximation for the error field $\hat{\phi}^*_e$, combined with the normal input field $\phi_a$. In case of losses it is beneficial to do both (as we show later). Practically this can also be performed in three measurements, as we can simply send in light from two directions at once and measure the sum of their intensities (if we average out the intensity fluctuations caused by standing waves).
Before we move to numerical results, it is worth considering what form the unitary transforms would take on a chip. We found that--as long as $\U_i$ sufficiently `mixes' the states--it's exact form matters very little for final performance. With sufficient mixing we mean that each exiting waveguide should contain light of a sufficiently large number of entering waveguides, such that information of each input channel is effectively spread over the chip. In the numerical experiments that follow we constructed the $\U_i$ by combining multiple 50/50 directional couplers (See Figure \ref{fig:chip}(b) for more details).
\subsubsection*{Numeric Simulations}
\begin{figure}[t]
\begin{center}
\includegraphics[width=0.6\textwidth]{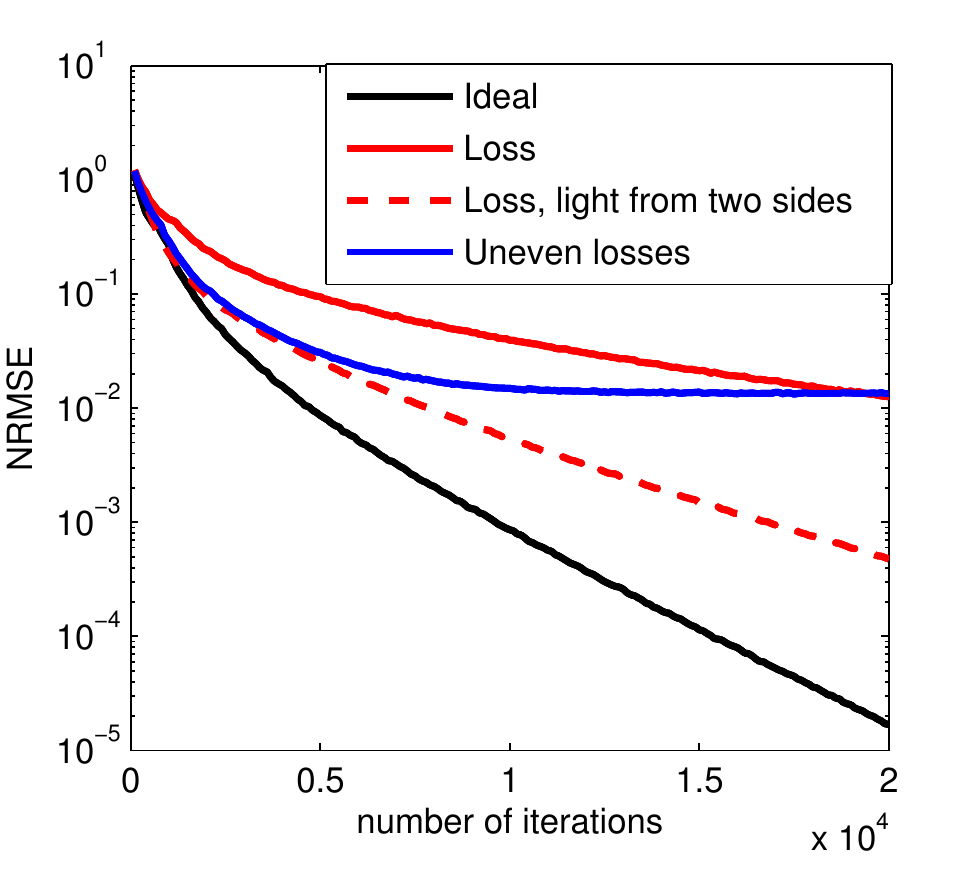}
\caption{Results of training the optical chip to emulate a pre-specified unitary transformation. Four different scenarios are considered, taking under consideration different degrees of non-ideal behaviour.}\label{fig:chip_results}
\end{center}
\end{figure}
We will simulate the training process of a chip with 50 inputs and 50 outputs. The details of the experiments are explained in the supplementary material. We test the following scenarios:
\begin{itemize}
\item First of all we consider the ideal, lossless scenario. 
\item Next we assume a 6\% power loss in each waveguide/phase shifter, based on assuming we couple out one percent of power for local measurements, and a 5\% intrinsic loss (which was the lowest number we could find in the integrated photonics literature \cite{Harris2014}). 
\item We assume the same losses as before, but now we compute two gradients resulting from the two directions light can travel through the chip, and we use the average gradient of that.
\item Finally we assume uneven losses, where each waveguide loses an additional fraction of power, uniformly and randomly picked between zero and one percent. Here we will use the average gradient again, resulting from two directional modes.
\end{itemize}
Figure \ref{fig:chip_results} shows the evolution of the NRMSE for each of the four experiments. Under ideal circumstances, the chip performs very well and the error goes down to negligible levels. When the chip is lossy, the training process is hampered. This is mostly due to the fact that power levels decrease by almost a factor 80 (-19 dB) from entrance to exit, which means that the gradients obtained for phase shifters close to the light source are far larger than those further away. Indeed, when we use light coming from both directions this problem is partially eliminated. Note that there are probably more elegant ways to avoid this problem, for instance by having different `sensitivities' for different phase shifters at different locations, such that internal losses are compensated completely.\\
Finally, uneven losses seem to pose the biggest challenge. One needs to be careful to interpret this result, however: due to uneven losses, the chip will be intrinsically unable to produce a unitary transformation from input to output, as the intermediate stages no longer are unitary. This means that the NRMSE will always have a certain lower bound given by the uneven losses. To check to what degree the result is due to this bound or to impairment in the training process, we simply re-measure the NRMSE after the training phase on the same (trained) chip, but without the uneven losses. It turns out this reduces the resulting NRMSE by a factor about 2 ($\textrm{NRMSE} \approx 0.007$). If we do not keep the learning rate fixed throughout training, but let it drop linearly to zero, this even becomes almost a factor 10 ($\textrm{NRMSE} \approx 0.0015$), while there is no difference in final performance for the chip \emph{with} the uneven losses. This shows that uneven losses do not seem to critically endanger the training process.\\
One final observation is that the trained phase shifts have a relatively small standard deviation. We initialised them at zero, and after training, we observed that nearly all the phase shifts are still within $\left[-2\pi/10,2\pi/10\right]$, i.e., only covering one fifth of the full $2\pi$ range. This is important as it implies that we do not necessarily need phase shifters which can cover this entire range. Indeed, truncating the phase shifts within the aforementioned range does not visibly affect performance in any of the presented experiments.
\section*{Discussion}
Many more factors need to be taken into account before a trainable MVM using waves may be constructed in practice. For the nanophotonic implementation, two important remaining factors we haven't yet studied are measurement noise and scalability. The first is concerned with how a noisy measurement will affect the obtained gradient. Indeed, we do not wish to lose a lot of power in the local power measurements within each waveguide/phase shifter. This means that the adaptations of the chip will need to be based on measurements of very low optical power, perhaps close to the noise floor. One factor that works in our advantage is the fact that these power measurements do not need to happen extremely fast. Whereas in typical telecommunication applications photodiodes need to measure optical power at rates well over the Gigahertz range, we can work with far lower measurement rates. Suppose for example we allow one microsecond for each of the three required power measurements, this would still allow several hundreds of thousand updates per second, while allowing a relatively long measurement time (i.e., time to accumulate energy) for each phase shifter. 
\\
Scalability in terms of allowable losses pose a more significant challenge. Right now we simulated a chip where the light needs to travel through 70 arrays of phase shifters, losing almost 99 \% of power in the process. Scaling the chip up to larger proportions will certainly exacerbate this problem, and one would very quickly reach a situation in which the chip would be unusable. This means that scalability of nanophotonic trainable chips would hinge completely on the ability of using low-loss components, or potentially on the use of optical amplification that compensates the losses. Here, again, one factor that plays in our advantage is the fact that the mechanism used for phase shifting within the chip doesn't need to have a fast response time. Note that in typical (e.g., telecom) applications, speed is more important than losses. This means that one of the most typical design constraints is lifted, and when developing suitable phase shifters for trainable optics, one can focus on compactness and low losses.
\\
The proposed physical optimisation method of this paper may also find applications outside of neural networks. It could be used more generally to automatically tune optical linear systems if a specified target transformation is not available, but when an error (a difference between a desired and actual output field) can be defined. This may have applications in decoding multimode communication channels (similar to \cite{Miller2013}), when the channel is not reliable (changing over time). A fast adaptation method based on physically implemented gradient descent may be useful here.

\noindent 

\section*{Acknowledgements}
MH has been supported by the Interuniversity Attraction Poles (IAP) program of the
Belgian Science Policy Office (BELSPO) (IAP P7-35 ``photonics@be''). TVV wants to thank the Special Research Fund of Ghent University (BOF) for financial support through a postdoctoral grant.

\section*{Supplementary Material Starts Here}

\section{Gradient descent on the Helmholtz equation}\label{section:helmholtz}
In this section we explain how we can find the gradient of a certain cost function w.r.t. a certain parameter for a system described by the Helmholtz equation. Suppose we have a field $\phi_a(\r)$ which adheres to the following equation:
\begin{eqnarray}
\nabla^2\phi_a(\r) + k^2(\r)\phi_a(\r) &= &a(\r)\textrm{ for } \r\in\Omega\label{eq:debc}\\
\phi_a(\r) &= &0 \textrm{ for } \r\in\partial\Omega,\label{eq:debc2}
\end{eqnarray}
where $\Omega$ is a closed domain of space, and $\partial\Omega$ is its edge, such that $\phi_a(\r)$ vanishes at the boundary. the function $a(\r)$ acts as the source, which in our case will encode the input vector of the MVM we wish to implement. We now introduce a certain cost functional $Q(\phi_a) $ that we want to minimize. In particular we wish to adapt the function $k(\r)$. Therefore, we are interested in finding the gradient of $Q(\phi_a)$ w.r.t. $k(\r)$. Normally, we can write:
\begin{equation}
 g(\r) = \frac{dQ(\phi_a)}{dk(\r)} = \int_\Omega\frac{\partial Q(\phi_a)}{\partial\phi_a(\r')}\frac{d\phi_a(\r')}{dk(\r)}d\r'.
\end{equation}
At this point however, care needs to be taken on how to interpret these derivatives. While $Q(\phi_a)$ and $k(\r)$ are both strictly real variables, $\phi_a(\r)$ is generally complex. The fact that $Q(\phi_a)$ is strictly real means that it is not an analytic function, and we need to write: $Q(\phi_a, \phi_a^*)$, the asterisk $*$ indicating the complex conjugate. We need to apply the chain rule separately for $\phi_a(\r)$ and $\phi_a^*(\r)$ (using the so-called Wirtinger derivatives). We find that
\begin{eqnarray*}
 g(\r) &=& \frac{dQ(\phi_a,\phi_a^*)}{dk(\r)} \\
 &=&\int_\Omega\frac{\partial Q(\phi_a,\phi_a^*)}{\partial\phi_a}\frac{d\phi_a}{dk(\r)}d\r' \\
 & +&\int_\Omega\frac{\partial Q(\phi_a,\phi_a^*)}{\partial\phi_a^*}\frac{d\phi_a^*}{dk(\r)}d\r'
\end{eqnarray*}
Since $Q(\phi_a,\phi_a^*)$ is strictly real, it follows that
\[
\frac{\partial Q(\phi_a,\phi_a^*)}{\partial\phi_a} = \left[\frac{\partial Q(\phi_a,\phi_a^*)}{\partial\phi_a^*}\right]^*.
\]
Similarly, we find that 
\[
\frac{d\phi_a}{dk(\r)} = \left[\frac{d\phi^*_a}{dk(\r)}\right]^*,
\]
such that we can rewrite the equation as:
\begin{equation}
 g(\r) = 2\Re\left(\int_\Omega\frac{\partial Q(\phi_a,\phi_a^*)}{\partial\phi_a(\r')}\frac{d\phi_a(\r')}{dk(\r)}d\r'\right).\label{eq:grad1}
\end{equation}
 We will use a more compact notation for the first factor:
\begin{equation}
 e(\r) = \frac{\partial Q(\phi_a,\phi_a^*)}{\partial \phi_a(\r)}.
 \end{equation}
The second factor can be found when we consider Equation \eqref{eq:debc} and \eqref{eq:debc2} as a linear operator $\mathcal{L}$, such that we can rewrite them as:
\begin{equation}
 \mathcal{L}\phi_a(\r) = s(\r),\label{eq:de_operator}
\end{equation}
where $\mathcal{L}\phi_a(\r) = \nabla^2\phi_a(\r) + k^2(\r)\phi_a(\r)$ and $s(\r) = a(\r)$ for $\r\in\Omega$. For points on the boundary $\r\in \partial\Omega$ we define $\mathcal{L}$ as the identity:  $\mathcal{L}\phi_a(\r) = \phi_a(\r)$, and $s(\r) = 0$, such that the condition of Equation \eqref{eq:debc2} is absorbed into the definition of $\mathcal{L}$, and Equation \eqref{eq:de_operator} describes both the Helmholtz equation and the boundary conditions. Taking the derivative of Equation \eqref{eq:de_operator} w.r.t. $k(\r)$ we find:
\begin{equation}
\mathcal{L}\frac{d\phi_a(\r')}{dk(\r)} = - \frac{\partial\mathcal{L}}{\partial k(\r)}\phi_a(\r').
\end{equation}
The right hand side is a functional derivative, and is equal to $-2k(\r)\phi_a(\r)\delta(\r - \r')$, with $\delta(\cdot)$ the dirac delta function. The left hand side is the operator $\mathcal{L}$ operating on $\frac{d\phi_a(\r')}{dk(\r)}$. Therefore, $\frac{d\phi_a(\r')}{dk(\r)}$ is the solution of the same Helmholtz equation with the same boundary conditions, but a different source function.\\
We will use the Green's function $G(\r,\r')$ that acts as the inverse of operator $\mathcal{L}$, which allows us to write:
\begin{equation}
 \phi_a(\r) = \int_\Omega G(\r,\r')s(\r')d\r',
\end{equation}
as a solution to $\mathcal{L}\phi_a(\r) = s(\r)$. When using $-\frac{\partial\mathcal{L}}{\partial k(\r)}\phi_a(\r')$ as source function, this leads to:
\begin{eqnarray}
 \frac{d\phi_a(\r')}{dk(\r)} &= &-\int_\Omega G(\r',\r'')\frac{\partial\mathcal{L}\phi_a(\r'')}{\partial k(\r)}d\r''\nonumber\\
 &=&  -2G(\r',\r)k(\r)\phi_a(\r).
\end{eqnarray}
Inserting this in Equation \eqref{eq:grad1} yields:
\begin{equation}
 g(\r) =  -4k(\r)\Re\left(\phi_a(\r)\int_\Omega G(\r',\r)e(\r')d\r'\right).
\end{equation}
Note that the arguments of the Green's function in the integral have switched places. It can be proven that the Green's function associated with Equation \eqref{eq:debc} is self-adjoint, which means that $G(\r,\r') = G(\r',\r)$. Therefore we can write:
\begin{equation}
 g(\r) = -4k(\r)\Re\left(\phi_a(\r)\int_\Omega {G(\r,\r')e(\r')}d\r'\right).
\end{equation}
In other words, if we define a second field which adheres to the equation $\mathcal{L}\phi_e(\r) = e(\r)$ this becomes:
\begin{equation}
 g(\r) = -4k(\r)\Re\left(\phi_a(\r)\phi_e(\r)\right).
\end{equation}
Both $\phi_a(\r)$ and $\phi_e(\r)$ can be obtained in a physical manner as they are solutions to the same wave equation with different source terms.\\
We can simplify the gradient even further when we assume that $k(\r)$ only varies slightly around a fixed average value which only has a global scaling effect on the gradient:
\begin{equation}g(\r) \sim -\Re(\phi_a(\r)\phi_e(\r)).\end{equation}

\section{Numerical example of an MVM}
We use a planar medium with ten sources / receivers, which are simulated as having 
\[\beta_i(\r) = \frac{\exp\left(-\frac{||\r - \r^\beta_i||^2}{2\sigma^2}\right)}{2\pi\sigma^2},\]
\[\gamma_i(\r) = \frac{\exp\left(-\frac{||\r - \r^\gamma_i||^2}{2\sigma^2}\right)}{2\pi\sigma^2},\]
with $\r^\beta_i$ and $\r^\gamma_i$ are the locations of the $i$-th source and receiver, respectively. The parameter $\sigma$ we chose to be approximately equal to half a wavelength in the medium surrounding the sources and receivers. The spacing between the sources and receivers was slightly over two wavelengths.\\
In between the sources and receivers is the medium of which we spatially modulate the wave number. Each iteration we draw a vector $\a_k$ and use this to generate the source $a(\r) = \sum_{i=1}^{N_a} a_i \beta_i(\r)$. We compute the complex field $\phi_a(\r)$, and the output $o_i = \int_\Omega \phi_a(\r)\gamma_i(\r) d\r$. Next, we redo the simulation with the error source term $e(\r)$, and compute the resulting complex field $\phi_e(\r)$. After both simulations, we compute the gradient and update $k(\r)$ in the region which was designated as the trainable medium.\\
We numerically solve the respective Helmholtz equations using the stabilised biconjugate gradient method. The simulation area is represented by a rectangular grid of elements, which have a size of roughly one tenth of the wavelength (for the average wave number of the system). To speed up the convergence of the computation of the complex fields we use an absorbing boundary layer, which we implement by adding a gradually increasing imaginary part to the wave number. Note that this is not a necessary element for the physical setup. We could for instance put the trainable medium between two mirrors, such that less energy is lost between the source and receivers.\\
We used 1000 training iterations. The matrix $\W$ was picked to be a random complex matrix, and was next scaled down with a factor 25. This scaling was necessary as only a fraction of the total energy of the sources is radiated onto the receivers. Interestingly, when we used much larger scaling factors for $\W$, the training algorithm adapts $k(\r)$ to extreme values in an attempt to reflect as much energy as possible back to the receivers (eventually leading to the inability to compute the complex fields). 
The learning rate $\eta_k$ we picked at a suitable starting value $\eta_0$, and we let it drop linearly to zero over the course of the training phase: $\eta_k = \eta_0(1 - k/N_\textrm{it})$, with $N_\textrm{it}$ the total number of iterations. As a qualitative measure of performance of the system at the $n$-th iteration we use the normalised root mean square error (NRMSE), equal to $||\o_n - \o^t_n||/\sqrt{\left<||\o_k^t||^2\right>_k}$, where $\left<\cdot\right>_k$ means the average over $k$.

\section{Nanophotonic implementation}
We will obtain gradients using the following expression:
\begin{eqnarray}
2\Re(\phi_a\phi_e) &=& \phi_a\phi_e + \phi_a^*\phi_e^*\nonumber \\
&=& (\phi_a + \phi_e^*)(\phi_a + \phi_e^*) - \phi_a\phi_a^* - \phi_e\phi_e^*\nonumber\\
&=&I(\phi_a + \phi_e^*) - I(\phi_a) - I(\phi_e)\label{eq:grad33},
\end{eqnarray}
which requires the measurement of three intensities, and the generation of the complex conjugate field of either $\phi_a$ or $\phi_e$.
\subsection{Error sources within waveguides}
Note that there is an important difference between the way light enters a photonic chip, and the theory presented in Section \ref{section:helmholtz}. Here, we assumed that there exists a source term, which for EM radiation would imply an optical antenna (represented in the maxwell equations by an alternating current with the same frequency as the light). In reality, however, light from an external laser source entering a chip is modulated in phase and amplitude, which means there is no internal `source' within the chip. This means that, when we generate the field $\phi_e$, we need to modulate the light at the point of measurement \emph{as if} it was produced by a source term. It turns out that  this means that we need to multiply the complex field of the `error' light entering the chip with a factor $-\jmath$. This can be proven when we apply the theory of  Section \ref{section:helmholtz} to a waveguide with guided modes. We start by writing the `output' field exiting a chip in the form of a guided mode in a waveguide as:
\begin{equation}
\phi_a(\r) = p(y)\exp(\jmath k_0x)\phi_o,\label{eq:guided}
\end{equation}
where $p(y)$ is the normalised transversal field profile of the guided mode, $k_0$ is the effective wave number of that particular guided mode, and $\phi_o$ is a complex variable that will be the effective output. Note that this implies that $y$ is the transversal direction, the waveguide lies according to the $x$-axis, and light is traveling in the positive $x$-direction. We assume that the output is generated at $x = 0$ by $\gamma(\r) = \delta(x)p(y)$, such that 
\[
o = \int_\Omega{\phi_{a}(\r)\gamma(\r) d\r} =\phi_o.
\]
Conversely, the source term for the error becomes 
\[
e(\r) = \frac{\partial Q}{\partial o}p(y)\delta(x).
\]
We assume $\phi_e$ can be split up into different coordinate parts as well, with the transversal part having the same field profile i.e.:
\[
\phi_e(\r) = q(x)p(y).
\]
When we insert this into the Helmholtz equation with the source term stated above, we obtain:
\begin{equation}
p(y)\frac{\partial^2 q(x)}{\partial x^2} + q(x)\frac{\partial^2 p(y)}{\partial y^2} + k^2(y)q(x)p(y) = \frac{\partial Q}{\partial o}\delta(x)p(y)\label{eq:pq_split}.
\end{equation}
Note that $k(\r) = k(y)$, as we assume a waveguide structure which doesn't change in the $x$-direction. We can eliminate $p(y)$ by inserting Equation \eqref{eq:guided} into the homogeneous Helmholtz equation. This leads to:
\begin{equation}
\frac{\partial^2 p(y)}{\partial y^2} = p(y)\left( k_0^2 - k^2(y)\right).\label{eq:p_solution}
\end{equation}
If we insert Equation \eqref{eq:p_solution} in Equation \eqref{eq:pq_split}, and divide by $p(y)$, we obtain:
\[
\frac{\partial^2 q(x)}{\partial x^2} + k_0^2 q(x) = \frac{\partial Q}{\partial o} \delta(x).
\]
The solution to this equation can be found by using the one-dimensional Green's function of the Helmholtz equation, which yields:
\[
q(x) = -\jmath\frac{\partial Q}{\partial o}\frac{\exp(\jmath k_0 |x| )}{2k_0}.
\]
We are only concerned with the light going back into the chip, i.e., for $x<0$, which indeed represents light propagating into the negative $x$-direction and therefore into the chip. In order to `simulate' a source with these properties, we need to modulate the incoming light with the error signal $\frac{\partial Q}{\partial o}$, times a factor $-\jmath$ (ignoring the scaling factor $(2k_0)^{-1}$).
\subsection{Abstraction of the chip}
When we describe the optical fields at the input waveguides with $\a$ and those at the output with $\o$, the transformation performed by the entire chip can be written as
\[
\o = \W\a = \left[\prod_{i=1}^N \P_i\U_i\right]\a,
\]
where $\P_i = \textrm{diag}(\exp(\jmath\boldsymbol{\psi}_i))$, a diagonal matrix with phase shifts written in vector form as $\boldsymbol{\psi}_i$, describing the phase shifts performed by the waveguides after the $i$-th unitary transformation of the signal. Concretely, after abstracting the operation of the chip we find that the complex field of the light before entering the $k+1$-th unitary transformation is given by:
\[
\a_k = \left[\prod_{i=1}^k \P_i\U_i\right]\a
\]
For the error light, entering from the opposite direction, the complex field of the incoming light at the same locations is given by:
\[
\e_k = \left[\prod_{i=0}^{N-k-1} \U^\textsf{T}_{N-i}\P_{N-i}\right]\e,
\]
with $\e =\e_N= -\jmath(\o - \o_t)^*$, the output error. We wish to apply Equation \eqref{eq:grad33}, so we will need to measure the complex state $\e_0$ at the point where the original input $\a$ was inserted, and re-emit its complex conjugate back into the network, added to the original $\a$. After all, when the system is completely unitary we can write:
\begin{equation}
\e_k^* = \left[\prod_{i=1}^k \P_i\U_i\right]\e_0^*.
\end{equation}
when $\U_i$ and $\P_i$ are lossy, however, this is no longer true, and we can only generate an approximation of the complex fields:
\begin{equation}
\hat{\e}_k^* = \left[\prod_{i=1}^k \P_i\U_i\right]\e_0^*.
\end{equation}
When we define $\h = \a + \e_0^*$, this field at the same locations throughout the chip becomes:
\[
\h_k = \left[\prod_{i=1}^k \P_i\U_i\right]\h
\]
We will denote the gradient for the phase shift $\boldsymbol{\psi}_i$ with $\g_i$. Following Equation \eqref{eq:grad33} we can write this gradient as:
\begin{equation}
-\g_i \sim \h_i\h_i^* - \a_i\a_i^* - \hat{\e}_i\hat{\e}_i^*\label{eq:grad_chip}
\end{equation}
which correspond to the intensities at the locations of the waveguides. Intensity measurements will cause a small loss everywhere, and typically variable phase shifters are lossy too. Let's incorporate this effect; we assume that $\P_i = \textrm{diag}(\lambda\exp(\jmath\boldsymbol{\psi}_i))$, with $\lambda$ a loss factor slightly smaller than one. In this case $\hat{\e}_k$ decays throughout the chip at the same rate as $\a_k$. In the end this means that, even though the signals lose power throughout the chip, the phase relations between $\a_k$ and $\hat{\e}_k$ remain correct. Indeed, in the numerical simulation we show that the resulting gradient is effective in training the conceptual optical chip.
\\
Note that we can also obtain the gradient using light going in the opposite direction by approximating the complex conjugate field $\phi_a$ as 
\[
\hat{\a}_k^* = \left[\prod_{i=0}^{N-k-1} \U^\textsf{T}_{N-i}\P_{N-i}\right]\o^*,
\]
which would give an alternative gradient:
\begin{equation}
-\g'_i \sim \h'_i{\h'_{i}}^{*} - \hat{\a}_i\hat{\a}_i^* - \e_i\e_i^*, \label{eq:grad_chip2}
\end{equation}
\[
\h'_k = \left[\prod_{i=0}^{N-k-1} \U^\textsf{T}_{N-i}\P_{N-i}\right]\h',
\]
and $\h' = \e + \o^*$. Indeed, if the chip is lossy it is desirable to take the average of both approximate gradients. This will even out the effects of losses and perhaps reduce approximation errors.

We will also simulate what happens when the transformations $\U_i$ aren't perfectly unitary. Suppose for instance that a certain amount of energy is lost, but a different amount for each output channel. Effectively, the transformation could be written as $\U_i = \D_i\U'_i$, with $\U'_i$ a true unitary matrix and $\D_i $ a diagonal matrix with elements smaller than one on the diagonal, which all differ from each other. This adds another problem to training the chips, as unevenly distributed losses will affect phase relations when attempting to create the field $\phi^*_e$. Of course, there are gradations to this issue. If the differences in losses are only slight, it might only affect the operation to a limited degree, and one would still obtain useful gradients throughout the chip.
\\
We will simulate the training process of a chip with 50 inputs and 50 outputs. To have a sufficient number of trainable parameters we assume $N = 70$ phase shifter arrays. It can be proven that a unitary matrix of size $M \times M$ is defined by $M^2$ parameters (as opposed to a random square complex matrix which has $2M^2$ parameters). It is unclear if the chip structure as we presented is theoretically able to emulate all possible $M \times M$ unitary matrices if it has only $N = 50$ layers of phase shifters (which would provide the right number of parameters). In practice we find that $N=50$ does not converge to a satisfactory level. More layers, lead to a much better performance. We settled for $N=70$ as a tradeoff between performance and additional losses.
\\
For each experiment we performed 20,000 training iterations, where as before, each iteration a single input / desired output pair was presented to the simulated chip. The learning rate $\eta$ we kept fixed during the training, but it was optimised in each of the four experiments to give the best final result. Note that the assumed losses will severely reduce the optical power that exits the chip. We want to train the chip to perform a unitary transform, however, which would preserve the total power. Therefore we `compensate' all losses by using only normalised input vectors and renormalizing the measurements at the output stages (both the output signal and the transformed error signal $\e_0$). These renormalised signals are then further used, both for training and for determining the NRMSE.
\end{document}